\documentclass[journal]{IEEEtran}

\usepackage{xcolor}
\usepackage{amsmath,amsfonts}
\usepackage{subcaption}
\usepackage{graphicx}
\usepackage{cite}
\usepackage{tabularx}
\usepackage{array} 
\usepackage{lipsum}

\usepackage[linesnumbered,ruled,vlined]{algorithm2e}
\usepackage{gensymb}
\usepackage{tabularx}
\usepackage{makecell}

\usepackage{graphicx} 

\title{\LARGE \bf Camera-RFID Fusion for Robust Asset Tracking in Forested Environments}

\begin{document}
\author{John Hateley, Sriram Narasimhan, Omid Abari%
\thanks{Hateley and Narasimhan are with the Department of Mechanical \& Aerospace Engineering at the University of California, Los Angeles and Abari is with the Department of Electrical and Computer Engineering at the University of California, Los Angeles (e-mail: jmhateley,snarasim@g.ucla.edu, omid@cs.ucla.edu).}%
}
\maketitle
\thispagestyle{empty}
\pagestyle{empty}
\begin{abstract}

Passive RFID tags offer a cost-effective and scalable solution for tracking numerous deployed assets. However, in forested environments, signal attenuation and multipath effects generally limit RFID spatial accuracy to the meter level. Conversely, while cameras employing stereo vision can achieve centimeter-level precision, relying solely on computer vision fails to resolve issues arising from spatial association ambiguity and partial occlusions in dense settings. Fusing these modalities allows systems to harness the high-accuracy benefits of vision while retaining the robust, non-line-of-sight identification advantages of RFID. Yet, a primary challenge in achieving this, which is the central focus of this paper, lies in accurately associating the disparate trajectories generated by these two sensors. To overcome this limitation, we introduce a novel camera--RFID fusion framework that integrates depth and object information with advanced trajectory-matching algorithms. By successfully bridging the meter-to-centimeter accuracy gap, the proposed approach helps achieve reliable tag localization even when assets temporarily leave the camera's field of view. To the best of our knowledge, this represents the first application of camera--RFID fusion for asset tracking in natural forested environments.

\end{abstract}
\section{Introduction}

Effective asset tracking is a critical aspect within the construction and forestry sectors—particularly in high-stakes operations such as timber harvesting and wildland firefighting—to enhance personnel safety, streamline workflow synchronization, and minimize operational costs. In densely forested or remote environments where terrestrial cellular infrastructure is fragmented or non-existent, localization traditionally relies upon Global Positioning Systems (GPS), Ultra-Wideband (UWB) sensors, or conventional radio frequency (RF) communication. However, the relatively high per-unit cost of GPS and UWB hardware often prevents comprehensive deployment across entire cohorts of personnel for many public agencies and small-scale private enterprises. Consequently, these organizations frequently revert to intermittent radio-based reporting, a method that introduces significant latency in spatial awareness and may lead to suboptimal outcomes, particularly during time-sensitive emergency situations.

Alternative tracking strategies, such as long-range passive RFID tags with mobile base stations, enable monitoring many assets at a fraction of the cost of GPS units. For outdoor asset tracking, we envision a cost-effective solution utilizing mobile platforms equipped with traditional localization capabilities such as GPS and readers tracking multiple assets tagged using passive RFID. In forested environments, robots can localize themselves to centimeter-level accuracy, with or without GPS. However, without cellular infrastructure, overall system localization typically achieves only 1–5 m accuracy, largely limited by the passive RFID components \cite{tag_10,tag_21}. Our recent work \cite{my_gp_paper} demonstrates that matching RF signal responses against a dictionary of environmental models can achieve GPS-level localization accuracy; however, these estimates remain subject to significant spatial uncertainty and low precision. Camera-based systems offer centimeter- to millimeter-level precision but often fail to distinguish visually similar assets or those obscured by objects, uniforms worn by personnel, or environmental factors common in forested environments \cite{camera_depth}.

\begin{figure}[htbp]
    \centering
    \small
    \includegraphics[width=1\linewidth]{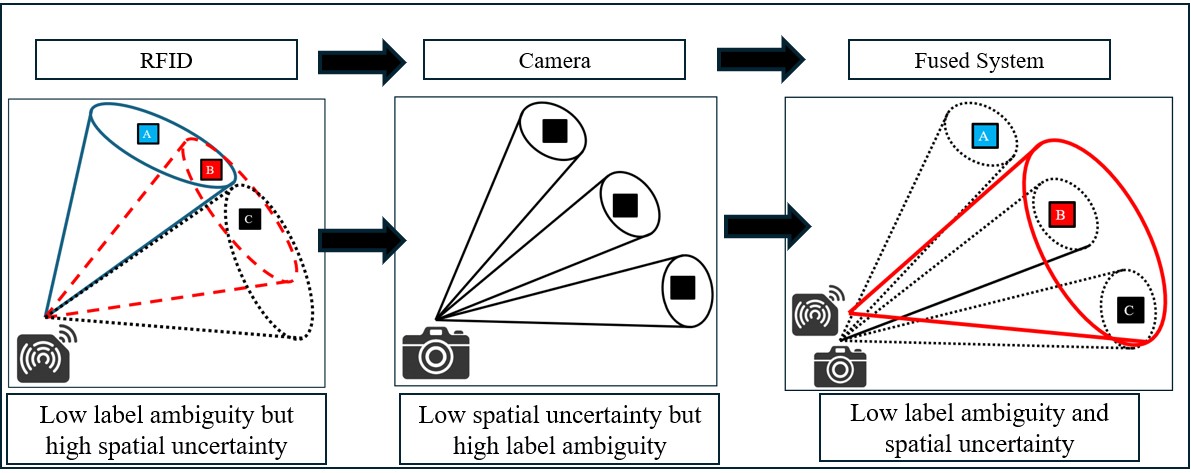}
    \caption{RFID systems are capable of unique identification but suffer from low positional accuracy in forest environments, while cameras can track accurately but cannot assign specific labels. The proposed solution fuses the two sensor modalities to provide an accurate labeled object with an accurate trajectory.}
    \label{Fig1_overallconcept}
\end{figure}


Integrating RFID and camera measurements is complementary in addressing their respective limitations and improving positioning accuracy \cite{camera_depth} (see Figure \ref{Fig1_overallconcept}). While cameras (such as depth or stereo) provide high spatial accuracy for detected objects, accurate object labeling---beyond generic classes such as people or vehicles---requires unique visual markers. In outdoor scenarios, many critical assets, such as loggers, construction crews or firefighters, often lack such visual markers or are temporarily occluded by debri or structures, making it difficult for cameras to distinguish between individuals or assets. By combining the high spatial accuracy of vision with the unique asset identifiers provided by RFID tags, reliable asset-specific tracking can be achieved during outdoor or forest based operations. This fusion also enables robust initialization of object positions for kinematic prediction when assets move outside the camera's field of view but remain within RFID range. Incorporating commodity RFID technology allows for scalability and cost-effectiveness according to specific scenarios and groups. However, such an integration introduces a crucial technical challenge of associating RFID measurements with the corresponding camera trajectories, which is a non-trivial gap in the current literature.

\textit{Contributions}: 
In this paper, we propose a novel trajectory-based fusion approach that combines the uncertain Fr\'echet distance with minimum-cost matching via the Mahalanobis distance to assign RFID tag identities to detected assets—a strategy that, to the best of our knowledge, is a first in camera–RFID fusion for localization. RFID tag trajectory is generated through a refined method (Gaussian Process regression) for range and angle of arrival prediction, while the trajectory of visually detected objects is obtained using depth measurements obtained from a commodity-grade stereo camera. Unlike existing fusion methods that operate at short ranges with well-separated tags, our system functions over relatively large distances (20–30 m) in realistic environments, where many tags in proximity to each other may be detected simultaneously and signal parameters alone are insufficient for reliable association. In addition, the increase in detection range and environment induced multipath effects create a disparity in error magnitudes between the sensor modalities, further challenging standard L2 Norm based trajectory matching schemes and making them unusable. We successfully demonstrate our approach in a real forested environment with up to four individuals in a 52 $m^2$ shared area.

\section{Related Work}

\subsection{RFID based Localization}
A passive RFID tracking system consists of an RFID reader and a passive RFID tag, where the tag is powered by the high-power RF signal transmitted by the reader. The tag reflects this signal back to the reader, appending its unique ID to the reflected transmission, a process commonly referred to as backscatter in the RFID literature. The reader can then extract information such as the reflected signal’s phase ($\phi$), frequency ($f$), time of reception, and received signal strength (RSSI) to estimate the tag’s position relative to the reader using methods such as RSSI-based ranging or phase difference measurements. Passive RFID tags have been employed in a wide range of localization tasks across both indoor and outdoor environments, including the tracking of mobile robots via landmark positions as discussed in \cite{acm1,tag_6,tag_10,tag_21}, and with robot interaction in visually occluded environments such as in \cite{icra_1}.

However, this approach has crucial weaknesses; RFID systems are highly susceptible to environmental obstructions such as trees, barriers, or adverse weather, which can distort signals and reduce localization accuracy \cite{tag_14}. To address these challenges, machine learning techniques such as environmental fingerprinting \cite{tag_2,tag_22}, convolutional neural networks \cite{tag_17,tag_24}, and most recently environmental modeling \cite{my_gp_paper},  have been utilized to train systems to recognize and compensate for such variations. Although these approaches can enhance performance in structured indoor or in well-trained outdoor environments, the spatial uncertainty in location estimates increases significantly when generalized to previously unseen settings or to large model dissimilarities, a significant limitation in dynamic scenarios in outdoor environmental monitoring. As a result, RFID systems in the localization context have largely been employed to serve as landmarks at known positions.    

\subsection{Trajectory Matching}
Trajectory matching and similarity analysis play a crucial role in object tracking, particularly in systems involving multiple motion-monitoring sensors. These techniques facilitate the fusion and interpretation of motion data for downstream analytical tasks. Among the most widely used methods are the Fr\'echet Distance, Edit Distance, Longest Common Subsequence (LCSS), Euclidean norm and Dynamic Time Warping (DTW), each of which estimates the similarity between trajectories by computing a relative distance metric based on point-wise or segment-wise comparisons between curves \cite{frechet_seminal,edit_seminal,lcs_seminal,l2_seminal,dtw_seminal}. 

Minimizing these distance metrics enables the identification of adjacent or similar paths, supporting applications ranging from object tracking with sparse GPS signals and local maps \cite{traject_sim_2}, to the monitoring of pedestrian flows and migration patterns \cite{traject_sim_9,traject_sim_1}. Among these techniques, the Fr\'echet Distance has emerged as the most prominent due to its emphasis on preserving the directionality and continuity of movement, which is especially advantageous in scenarios where the temporal order and geometric shape of the trajectory are of primary concern. This characteristic has led to its widespread adoption in analyzing sparse datasets, as demonstrated in trajectory planning and movement reconstruction applications \cite{traj_app_1,traj_app_2}.

\begin{figure}[htbp]
    \centering
    \includegraphics[width=1\linewidth]{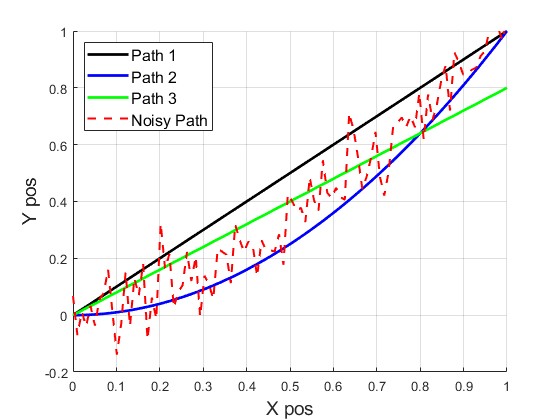}
    \caption{A example of a noisy trajectory overlayed with several trajectory candidates.}
    \label{noise_match_example}
\end{figure}

Recent work has examined trajectory matching when one or both trajectories contain significant noise, as shown in Fig. \ref{noise_match_example}. Trajectory matching fundamentally relies upon the spatial proximity of points along two curves as their proximity indicates the degree of matching \cite{frechet_seminal,edit_seminal,lcs_seminal,dtw_seminal}. Under noisy conditions, trajectory matching becomes challenging because high uncertainty degrades position accuracy. To address this, studies have applied the Fr\'echet Distance alone or combined it with metrics such as the Manhattan Distance or Wasserstein Metric to compare trajectories \cite{traject_sim_3,traject_sim_4,traject_sim_7,traject_sim_10}. These methods model uncertainty by creating probability distributions around noisy data points and evaluating the minimum and maximum distances between candidate paths, followed by statistical analysis to assess trajectory similarity.

\subsection{Camera-RFID Association}\label{traject:cam_rfid}

Cameras and RFID can be complementary to each other in their capabilities: RFID provides persistent identity and works under non line of sight or poor visibility conditions, while cameras give precise positioning and motion details under ideal conditions and at close ranges. A hybrid approach fusing their complementary capabilities is especially valuable in dynamic outdoor tracking of assets (human or non-human assets)—where identity accuracy and spatial precision are both critical. For example, \cite{cam_rfid_1} combined passive RFID tags with a monocular camera to identify books by associating the timestamps of RFID detections with the camera’s object recognition outputs. Similarly, studies such as \cite{cam_rfid_6,cam_rfid_8,cam_rfid_10,cam_rfid_15,cam_rfid_14} employed RFID tags to track mobile assets as they moved across, into, or out of camera fields of view, correlating RSSI measurements with visually detected object positions. The resulting trajectory data was then used to reliably label each target object. This approach to matching was utilized again in \cite{cam_rfid_13}, where RFID tags were attached to marked objects and their trajectories were fused with the camera frame sequences to associate tag positions with specific objects. A similar approach was also used in \cite{cam_rfid_16} and \cite{icra_2} where the trajectory of RFID tags were fused with pixel positions to determine what objects in a frame need to be tracked or identified. However, these studies operate under constrained conditions where RFID tags can often be isolated so that only a single tag is detected at any time, as demonstrated in \cite{cam_rfid_1}, or can achieve centimeter-level localization accuracy comparable to that of camera-based systems \cite{tag_10,camera_depth,cam_rfid_13}. This combination of isolation and low localization error in these studies enables reliable object–tag association in these scenarios As far as we are aware, there are no prior studies that examine uncontrolled outdoor forested settings with overlapping tags and relatively large ranges on the order of 30 m, employing commodity passive RFID systems together with low-cost depth cameras.

\section{Methodology}

The overall methodology proposed in this paper consists of three main steps: (i) trajectory generation using the camera via computer vision and the RFID tags through a Gaussian Process model augmented Kalman filter to determine a tag's range and the signal's angle of arrival; (ii) estimating the Fr\'echet distance between the tag and object trajectories; and (iii) trajectory association through Mahalanobis Distance and minimum cost matching. The procedure to generate the camera trajectory through computer vision is described in Appendix A.

\subsection{Trajectory Generation Via Gaussian Process Augmented Kalman Filter} \label{trajec_gen}

The RFID range and angle of arrival (AoA) measurements are estimated via a trained Gaussian Process (GP) model, as raw phase measurements are heavily corrupted by multipath effects \cite{my_gp_paper}. To mitigate this noise and produce refined tag trajectories, a modified Extended Kalman filter (EKF) is implemented to estimate the tag state based on these GP-derived observations \cite{kalman,ekf}.

The state vector of the $i$-th tag at time $t$, denoted as $T_i(t)$, is defined in the cartesian coordinates:
\begin{equation}
T_i(t) = \begin{bmatrix} x_i(t) & y_i(t) & v_{x_i}(t) & v_{t_i}(t) \end{bmatrix}^T
\end{equation}
where $x_i(t)$ and $y_i(t)$ represent the positional estimate of the object, and $v_{x_i}(t)$ and $v_{y_i}(t)$ represent the corresponding velocities relative to the reader (mounted on a mobile platform).

The prediction step of the EKF utilizes a constant velocity (CV) model, and the state transition matrix, $\Phi$, projects the state forward, while the uncertainty growth is propagated through Equation \ref{uncertainty_prop}. Here, $Q$ represents the process noise covariance and is tuned to achieve filter stability \cite{kalman}:
\begin{equation}
    M_{i|i-1} = \Phi M_{i-1} \Phi^T + Q.
    \label{uncertainty_prop}
\end{equation}

To incorporate sensor measurements, the Cartesian state must be mapped into polar coordinates via the nonlinear observation function $h(T_i)$,
\begin{equation}
    h(T_i) = \begin{bmatrix}
        \sqrt{x_i(t)^2 + y_i(t)^2} \\
        \operatorname{atan2}(y_i(t), x_i(t))
    \end{bmatrix}.
\end{equation}
This transformation generates a predicted measurement, $z_{pred}$, used to calculate the innovation residuals during the update step. Since $h(T_i)$ is nonlinear, it is linearized via the Jacobian matrix $H_j$:
\begin{equation}
    H_j = \begin{bmatrix}
        \frac{x_i(t)}{\rho} & \frac{y_i(t)}{\rho} & 0 & 0 \\
        -\frac{y_i(t)}{\rho^2} & \frac{x_i(t)}{\rho^2} & 0 & 0 
    \end{bmatrix}
\end{equation}
where $\rho = \sqrt{x_i(t)^2 + y_i(t)^2}$ represents the radial distance. This matrix represents the local slope of the coordinate transformation, describing how polar observations vary relative to changes in the Cartesian state space.

The GP-derived measurements are integrated into the observation space using a predictive mean $f^*$ for a test point $x^*$, calculated as:
\begin{equation}
    f^* = m(x^*) + \mathcal{K}(x^*,X)\mathcal{K}(X,X)^{-1}(Y - m(X)) 
    \label{predic_mean}
\end{equation}
where $m(x)$ is the prior mean function, $Y$ represents the training ground truth for the set $X$, and $\mathcal{K}$ denotes the Radial Basis Function (RBF) kernel \cite{Rasmussen_Williams_2006}:
\begin{equation}
    \mathcal{K}(x_1,x_2) = \exp \left(-\frac{\lVert x_1 - x_2 \rVert^2}{2\sigma^2}\right)
\end{equation}

The corresponding predictive variance, $\sigma^2(x^*)$, provides a localized measure of uncertainty for each measurement:
\begin{equation}
    \sigma^2(x^*) = \mathcal{K}(x^*, x^*) - \mathcal{K}(x^*, X) \left[ \mathcal{K}(X, X) + \sigma_n^2 I \right]^{-1} \mathcal{K}(X, x^*)
    \label{gp_variance}
\end{equation}
where $\sigma_n$ is a small perturbation to ensure the invertibility of $\mathcal{K}$.

The resulting variances, $\sigma^2_r$ and $\sigma^2_\theta$, populate the measurement noise covariance matrix $V$. To initialize the state covariance matrix $M_0$, velocity variances are heuristically derived from the spatial uncertainty and the sampling interval $\Delta t$:
\begin{equation}
    \sigma_{v_x} = \frac{\sigma_x}{\Delta t}, \quad \sigma_{v_y} = \frac{\sigma_y}{\Delta t}
\end{equation}

The state covariance $M_i$ and the posterior state estimate $T_{i|i}$ are updated through the following recursive equations:
\begin{equation}
    \begin{aligned}
        S &= H_j M_{i|i-1}H_j^T + V \\
        K &= M_{i|i-1}H_j^TS^{-1} \\
        T_{i|i} &= T_{i|i-1} + K(z - z_{pred}) \\
        M_{i|i} &= (I-KH_j)M_{i|i-1}.
    \end{aligned}
\end{equation}

While the filter yields estimates for tag velocities, these latent variables are often unreliable due to environmental interference and the lack of direct velocity observations. As a result, the velocity components primarily serve to maintain the integrity of the smoothing process, while only the refined position estimates are passed to the subsequent trajectory association scheme.

\subsection{Trajectory Matching via Fr\'echet Distance}

In this problem, $N$ objects and $N$ RFID tags are detected by the camera, with the association between each tag and each visually detected object initially unknown. The camera and RFID trajectories are denoted as \( X_C \) and \( X_R \), respectively, in (\ref{trajectories}), with the camera trajectories converted to polar coordinates.

\begin{equation}
\begin{split}
X_{C_i} &= \{(r_{1_c}, \theta_{1_C}), (r_{2_C}, \theta_{2_C}), \dots, (r_{n_C}, \theta_{n_C})\} \\
X_{R_i} &= \{(r_{1_R}, \theta_{1_R}), (r_{2_R}, \theta_{2_R}), \dots, (r_{m_R}, \theta_{m_R})\}
\end{split}
\label{trajectories}   
\end{equation}

\subsubsection{Fr\'echet Distance}

The Fr\'echet Distance is trajectory matching scheme that relies upon the concept that trajectories are continuous and non-decreasing (i.e., no backtracking). This preserves the temporal order of points which are referenced in the same coordinate system. In discrete systems with set time stamps, the Fr\'chet Distance can be expressed as (\ref{discrete_frech_eq}), with $\sigma_1$ and $\sigma_2$ representing the order preserving traversals of the data sequence \cite{discrete_frechet_seminal}.

\begin{equation}
d_{dF}(X,Y) = \min_{\sigma} \max_{i=1,\dots,k} \| X_{C_{\sigma_1(i)}} - X_{R_{\sigma_2(i)}} \|
\label{discrete_frech_eq}
\end{equation}

To generate $X_{C_{\sigma_1(i)}}$ and $X_{R_{\sigma_2(i)}}$ in this paper, the two trajectories were first temporally aligned. The analysis interval used for alignment is defined as:

\begin{equation}
\begin{split}
t_{\text{start}} &= \max\left(t_{\text{start}}^{\text{cam}}, \, t_{\text{start}}^{\text{RFID}}\right) \\
t_{\text{end}} &= \min\left(t_{\text{end}}^{\text{cam}}, \, t_{\text{end}}^{\text{RFID}}\right)
\end{split}
\label{times}   
\end{equation}

After temporal alignment, spline interpolation was applied to both curves to sample the same number of data points between the two trajectories, enabling a consistent computation of the discrete Fr\'echet distance. 

However, the inherent noise in the system can lead to ambiguous trajectories and incorrect object association. This problem can be alleviated through the application of uncertain Fr\'echet Distance estimation \cite{traject_sim_3}:

\begin{equation}
\begin{split}
d_F^{\min} (X_C,X_R) &= \min_{x_c' \in X_C,x_R' \in X_R} d_F(x_c',x_R') \\
d_F^{\max} (X_C,X_R) &= \max_{x_c' \in X_C,x_R' \in X_R} d_F(x_c',x_R') \\
\end{split}
\label{uncertain_frechet}
\end{equation}

To account for the inherent noise in the system, the uncertain Fr\'echet Distance generates $K$ realizations for both modalities \cite{traject_sim_3}. Depth cameras exhibit centimeter-level accuracy compared to the meter-level ranging error of the RFID system, we treat the camera measurements ($X_c$) as trajectory ground truth. To support this assumption, note that the measured RMS depth noise is at the sub-pixel level \cite{keselman2017intel}, which translates to millimeter to centimeter-level real-world depth measurement error, i.e., two orders of magnitude smaller than typical RFID errors, even up to three orders of magnitude lower at close ranges.  This simplification enables us to reduce the computation time of the algorithm by only creating perturbations around the reader data which limits the amount of trajectories that have to be compared.

However, in this application the uncertainty region of each point $i$ cannot always be presented as core linear offset as covariances between the range and angle appear as the trajectory unfolds through the Kalman Filter. As a result, instead of simple linear offsets, the uncertainty region at each point $i$ is defined as an elliptical boundary derived from the Gaussian Process range and angle of arrival variances, $\sigma_\theta^2$ and $\sigma_r^2$  respectively \cite{traject_sim_3}. This boundary represents the confidence region of the tag's possible locations. The $K$ realizations are then sampled from this elliptical boundary, thus generating a set of candidate trajectories $x_R'$\cite{traject_sim_3}. This set is defined as:

\begin{equation}
x_R'[i] = \{ \mathbf{p} \in \mathbb{R}^2 \mid (\mathbf{p} - \hat{\mathbf{p}}_i)^T \Sigma_{i}^{-1} (\mathbf{p} - \hat{\mathbf{p}}_i) \leq \kappa^2 \}
\label{elliptical_uncertainty}
\end{equation}

where $\hat{\mathbf{p}}_i$ is the estimated polar position of the tag at time $i$, $\Sigma_{i}$ is the covariance matrix incorporating the predicted radial and angular variances, and $\kappa^2$ is the threshold for the desired confidence interval\cite{traject_sim_3}.

Once the new series of trajectories are generated, Equation \ref{discrete_frech_eq} is used with $x_R'$ and $X_C$ to compute the range of similarity values\cite{traject_sim_3}. This yields $d_{\min}$, representing the best-case alignment between the uncertain trajectories, and $d_{\max}$, capturing the worst-case divergence under RFID localization uncertainty. Since these values only provide a range of potential Fr\'echet Distances, the Mahalanobis Distance is subsequently employed to evaluate the distribution of $[d_{\min}, d_{\max}]^T$ for tag-object association\cite{traject_sim_3}.

\subsection{Tag-Object Association via Mahalanobis Distance and Minimum Cost Matching}

Due to the real time (order of seconds or minutes) needs often required in this application, it is possible to measure how the tag's trajectory aligns with the trajectories of various objects by including each new tag and object position in the analysis. As a result, for each detected object, we create a probability distribution ($N(\mathbb{E}[d_{\min},d_{\max}]^T,\Sigma)$) that models the proximity between the tag's trajectory and the trajectory of each detected object throughout a time interval $\mathcal{T}$. These different distributions are then compared using the Mahalanobis Distance \cite{mahalanobis_seminal}, defined as:

\begin{equation}
D_M(x(t),\mu) = \sqrt{(x(t)-\mu)^T\Sigma^{-1}(x(t)-\mu)}, \quad 0 \leq t \leq \mathcal{T}
    \label{mahalanobis}
\end{equation}

This equation measures the distance between a point $x$ and a probability distribution $N(\mu,\Sigma)$. In this application, $x = [0,0]^T$, which assumes a perfect match between the tag's trajectory and each of the detected objects, as shown below, where $E = \mathbb{E}[d_{\min},d_{\max}]^T$: 

\begin{equation}
D[i](X_C[i],x_R') = \sqrt{ E \space \space \Sigma^{-1} E ^T}
\label{modified_mahalanobis}
\end{equation}.

Applying Equation \ref{modified_mahalanobis} to each detected object produces a vector $M_R[i] = [D_1, D_2, \dots, D_N]^T$ which quantifies the similarity between the $i$-th tag’s trajectory and the trajectories of all detected objects. Aggregating these vectors into a matrix $M_R = [M_{R_1}, M_{R_2}, \dots, M_{R_N}]$ results in a matrix containing the relative proximity between all tag trajectories and object trajectories, with smaller values indicating closer alignment between the pairs.

Once the matrix $M_R$ has been constructed, each column represents how close a specific tag trajectory is to all the detected objects' trajectories, and each row corresponds to the trajectory of a particular object. To assign tags, each row is examined to find the tag with the smallest Mahalanobis distance, indicating the closest trajectory match. After a tag is assigned to an object, its corresponding column is removed from the matrix, and this process is repeated until every object has been paired with a unique RFID tag.

\section{Experiments} \label{traject:experiments}
This section describes the details on how the experiments (simulation and field) were set up, the data collection process, hardware and software used in the experiments, generation process for the trajectories, and baseline comparisons used to evaluate the proposed method. 

\subsection{Simulation Test Bed}

First, a simulation testbed was developed utilizing realistic operational ranges of commercially available RFID and vision-based subsystems (such as the ones used for field experiments in this study). A fundamental assumption of continuous tracking was maintained, wherein all visually detected entities were consistently monitored and associated for the duration of the experiment, a task that has been accomplished in real world experiments through software like Deep Sort \cite{deep_Sort}. The simulated agents, or ``targets,'' were tasked with navigating between discrete points on the test bed to simulate general movements. The primary metric of interest was the \textit{observation time} required for definitive tag-to-object association. For each trial, the simulation enforced a criterion of three consecutive correct associations to validate the result, with a temporal upper bound of one hour per trial to account for potential non-convergence.

\begin{table}[ht]
    \centering
    \footnotesize 
    \setlength{\tabcolsep}{2pt} 
    \caption{Maximum number of tags as a function of monitoring radius and occupancy density}
    \label{tags_per_region}
    \begin{tabularx}{\columnwidth}{|c|*5{>{\centering\arraybackslash}X|}}
    \hline
        Radius (m) & $8.0 \times 10^{-4}$ $p/m^2$ & $15.9 \times 10^{-4}$ $p/m^2$ & $23.9 \times 10^{-4}$ $p/m^2$ & $31.8 \times 10^{-4}$ $p/m^2$ & $39.8 \times 10^{-4}$ $p/m^2$ \\
         \hline
        20 & 1  & 2  & 3  & 4 & 5\\
         \hline
        30 & 3  & 5  & 7  & 9 & 12 \\
         \hline
        40 & 4  & 8  & 12 & 16 & 20 \\
         \hline
        50 & 7  & 13 & 19 & 25 & 32 \\
         \hline
        60 & 9  & 18 & 27 & 36 & 45 \\
         \hline
        70 & 13 & 25 & 37 & 49 & 62 \\
         \hline
        80 & 16 & 32 & 48 & 64 & 80 \\
        \hline
    \end{tabularx}
\end{table}

The experimental protocol consisted of 100 independent trials utilizing a variable density approach, with the number of tagged objects ($n$) ranging from 1 to 5 (tagged objects assumed to be persons carrying tags in our simulations). All objects were constrained to move within the sensor's assumed maximum range of 20~m, resulting in occupancy densities between $8.3 \times 10^{-4}$ and $39.8 \times 10^{-4} \text{ persons/m}^2$. The upper density threshold was established based on physical constraints, such as the collision risk posed by moving people in close proximity. The specific tag counts for each simulation and their corresponding test bed radii are summarized in Table~\ref{tags_per_region}.

\subsection{Hardware For Field Experiments}
The equipment used consisted of Zed 2 depth camera, Impinj R420 reader (which modulates the RF signal between 902.75 and 926.75 MHz), Ex0 3000 Passive RFID tags, 3 Vulcan RFID VUL-262006/TRH/A/K (RHCP) Outdoor RFID Antennas, and a HP Z2 Mini G9 Workstation. The Zed 2 camera utilizes the Zed SDK and a proprietary object detection algorithm to detect and localize the participants (bounding box level) during the experiments. For this study this built-in bounding box object detector was deemed adequate. 

When physically deployed, the RFID reader was only able to reliably detect off-center tags up to a distance of 20~m and within a maximum angle of $\pm 30^{\circ}$, with the effective range decreasing rapidly for every 15$^{\circ}$ from center. Additionally, the depth camera only had an effective object detection range of approximately 10~m. As a result of these sensing constraints and the confined deployment environment, only five to six participants could realistically operate within the shared detection region. This number was further limited by the performance of the person-labeling software, Deep SORT~\cite{deep_Sort}. Several labeling softwares were implemented during these scenarios to increase the accuracy of labeling between frames, but Deep SORT was the most reliable. Despite its reliability to detect multiple objects when tested in lab settings, in the field its performance degraded when there was a high density of individuals, leading to frequent identity switches and tracking failures. Consequently, the combined limitations of the sensing hardware and the labeling algorithm restricted the maximum number of reliably tracked participants in the physical deployment to four individuals. 

\subsection{Field Data Collection Process}

Data were collected in forested environments representative of typical West Coast forested locations. Eight RFID tags were placed at angles ranging from $\pm$30$^\circ$ and moved in 0.5 m increments from 2 m to 20 m from the base station. Calibration data were obtained by having volunteers hold RFID tags while both RFID and camera systems recorded measurements. Ten trials were conducted, each with 1–5 participants at different positions.

\begin{figure}[htbp]
    \centering
    \includegraphics[width=1\linewidth]{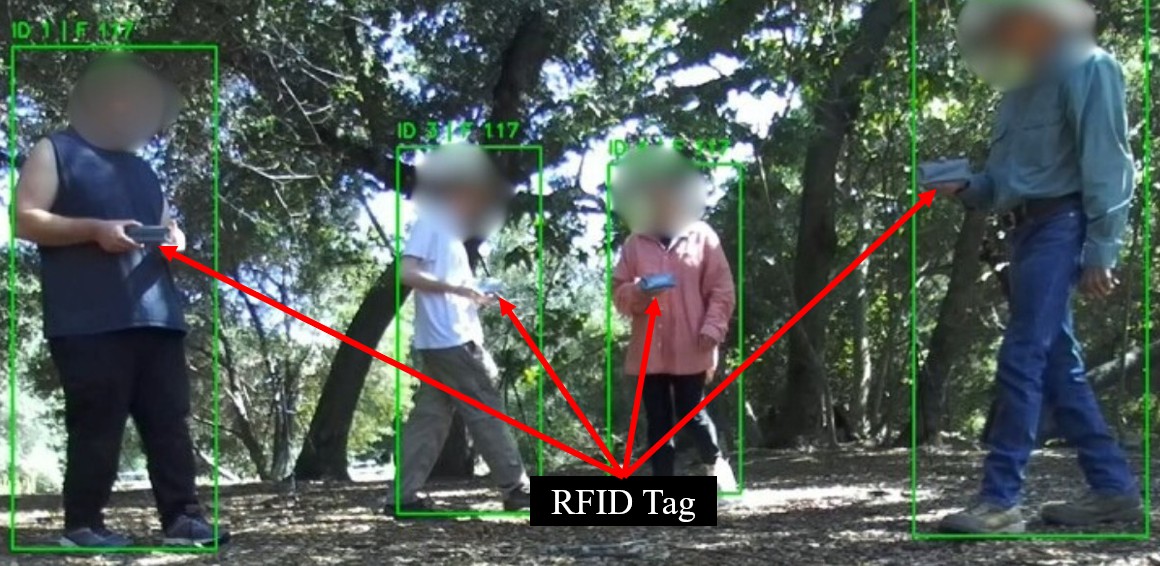}
    \caption{Four participants moving in the test environment while each holding an RFID tag; bounding box detections tracking individual participants also shown.}
    \label{fourpeople}
\end{figure}

Due to the Zed 2 camera’s 10 m detection limit, experiments were constrained to a 52 $m^2$ overlap region between the camera and RFID coverage areas, located 10 m from the base station with a 30$^\circ$ offset from the reader’s central axis. Empirically, five stationary participants filled the camera’s field of view without excessive crowding, but five moving participants led to frequent collisions and unrealistic behavior. Therefore, evaluation was limited to four participants (Fig. \ref{fourpeople}).

Trajectory data were collected by having volunteers move randomly for 3 min while carrying RFID tags, with each trial adding one additional participant up to four. Although each trial lasted 3 min, destructive interference at certain frequencies and FCC-mandated frequency cycling restricted usable localization data to the interval $[10,130]$ s.

\subsection{Trajectory Generation}

The camera employed in this experiment integrates a built-in person detection model at the bounding-box level, enabling the extraction of individual positions within each frame. To maintain label consistency across frames and to generate trajectories, we applied the Deep SORT algorithm \cite{deep_Sort}, which leverages object trajectories in conjunction with a Kalman filter. For high densities, we further incorporated a simple kinematic constraint that associated newly detected labels with previously tracked objects based on the Euclidean norm. This labeling strategy allowed us to reliably extract trajectories for all detected individuals.

\begin{figure}
    \centering
    \includegraphics[width=1\linewidth]{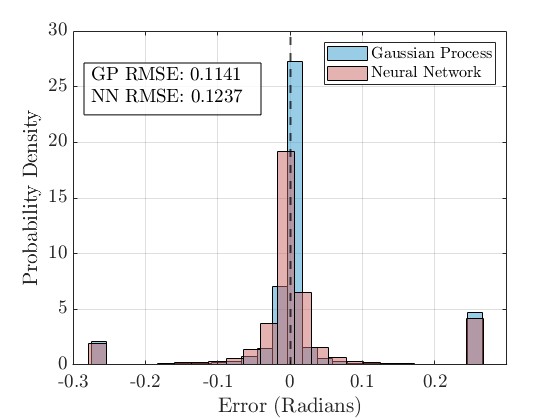}
    \caption{Comparison of Gaussian Process Models vs Neural Networks for angle of arrival estimation. The GP Models had a higher level of accuracy than optimized traditional neural networks to its ability to parse sparse data sets during predictions.}
    \label{gp_nn_comparison}
\end{figure}

To generate trajectories for the RFID tags, both the distance and angle of arrival relative to the antenna array are required. Both the distances and angle of arrival were estimated using the Gaussian Process (GP) models described in Section \ref{trajec_gen}. This machine learning framework is shown to outperform traditional Neural Network based estimation as seen in Figure \ref{gp_nn_comparison} (see this reference \cite{my_gp_paper} for details). Given that the reader operates across multiple frequencies, a distinct GP model was constructed for discretized phase bins, with each model being trained and evaluated on an 80-20 split.
\subsection{Baseline Comparisons}
This section presents methods used to compare the approach in this paper to fuse RF data with visual tracks with methods published in the literature. Two methods available in the literature include Euclidean Norm and Dynamic Time Warping (DTW) \cite{cam_rfid_2,cam_rfid_9,cam_rfid_13,cam_rfid_14,cam_rfid_16}. While both methods utilize distance metrics to evaluate similarity, they differ fundamentally in how they handle temporal alignment. The Euclidean norm approach calculates the $L_2$ norm by comparing corresponding points on two trajectories at each discrete time step, as shown in Equation \ref{l2norm_eq}:

\begin{equation}
    d(A,B) = \sqrt{\sum^n_{i=1}(a_i - b_i)^2}
    \label{l2norm_eq}
\end{equation}

where $d(A,B)$ represents the total distance between trajectories $A$ and $B$, while $a_i$ and $b_i$ are synchronized samples from each sequence \cite{l2_seminal}. This method requires both trajectories to be of equal length and perfectly aligned in time.

In contrast, Dynamic Time Warping (DTW) provides a flexible alignment by calculating distances between points regardless of their temporal index \cite{dtw_seminal}. The local cost between any two points $a_i$ and $b_j$ is typically defined by the squared Euclidean distance, as seen in Equation \ref{dis_pt_bet_pts}:

\begin{equation}
    \delta(a_i,b_j) = (a_i - b_j)^2
    \label{dis_pt_bet_pts}
\end{equation}

Unlike the Euclidean norm, which strictly compares points where $i=j$, DTW constructs a cumulative cost matrix to find the optimal warping path between sequences that may have different lengths or varying speeds \cite{dtw_seminal}. The recursive alignment process is defined in Equation \ref{dtw_eq}:

\begin{equation}
D(i, j) = \delta(a_i,b_j) + \min \begin{cases} 
D(i-1, j) & \text{(Expansion)} \\ 
D(i, j-1) & \text{(Contraction)} \\ 
D(i-1, j-1) & \text{(Match)} 
\end{cases}
\label{dtw_eq}
\end{equation}

The final value $D(n, m)$ represents the minimum cumulative distance, serving as the similarity metric between the two trajectories.

\section{Results and Discussion}

\subsection{Simulation Study}
\begin{figure}[hbtp]
    \centering
    \includegraphics[width=1\linewidth]{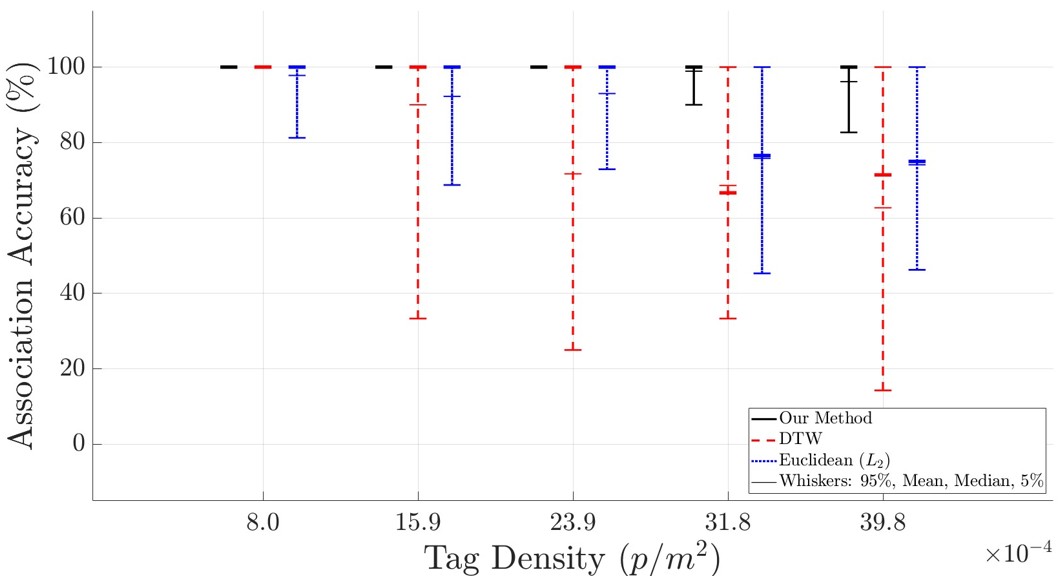}
    \caption{Association accuracy across 700 trials per density level. While the majority of the simulations achieved 100\% accuracy, the increased prevalence of linear trajectories in larger test beds led to a slight increase in variance at higher densities, with a global minimum accuracy of approximately 80\%.}
    \label{tag_density_plot}
\end{figure}

To evaluate the framework, the analysis focused on tag density within a defined monitoring region. Defining the operational envelope began with maximum detection range $R = 20~m$ to $ R = 80~m$, corresponding based on commercially available systems. The primary performance metric was the successful association rate between the camera and RFID trajectories over time across the trials conducted for a particular density (each column in Table \ref{tags_per_region}). Fig \ref{tag_density_plot} shows the results of the proposed method compared to DTW and the Euclidean Norm trajectory matching schemes.

\begin{figure}[htbp]
     \centering
     \begin{subfigure}[b]{0.45\textwidth}
         \centering
         \includegraphics[width=\linewidth]{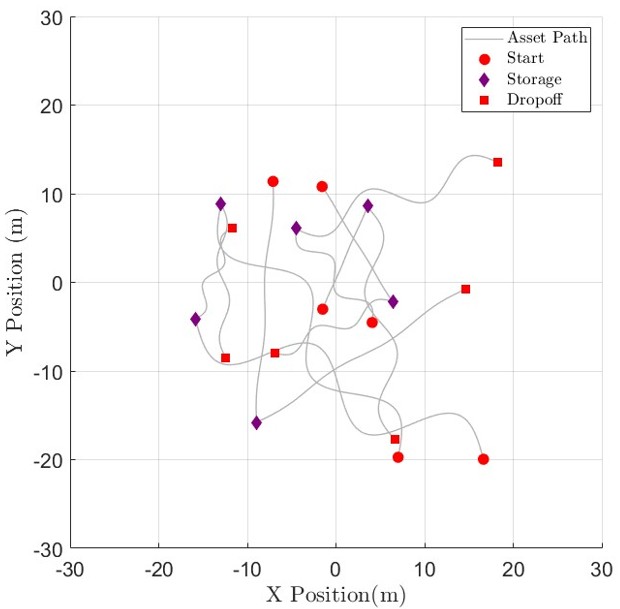}
         \caption{Tag Trajectories when $R = 20m$}
         \label{low_tag}
     \end{subfigure}
     \hfill 
     \begin{subfigure}[b]{0.45\textwidth}
         \centering
         \includegraphics[width=\linewidth]{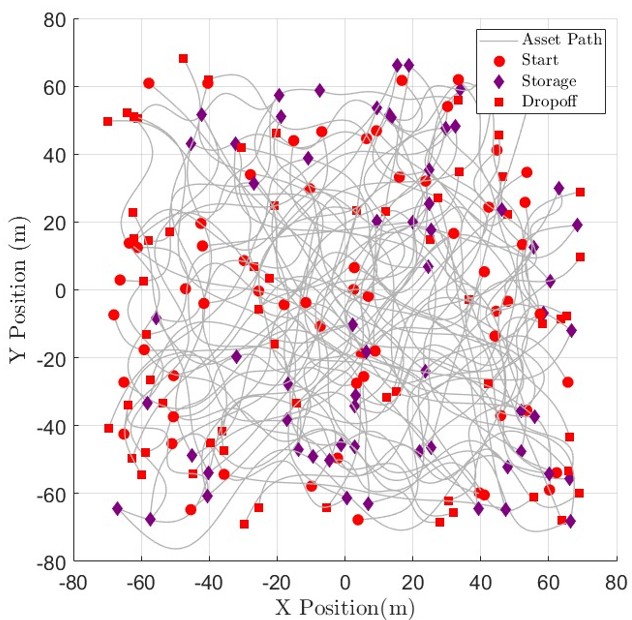}
         \caption{Tag Trajectories when $R = 80m$}
         \label{high_tag}
     \end{subfigure}
     
     \caption{Although density was constant the amount of tags present increased, thus increasing spatial overlap, assets with similar trajectories to one another which increased association difficulty.}
     \label{density_trajectories}
\end{figure}

Across $3,500$ trials ($100$ trials per tag count), accuracy typically fluctuated between $80\%$ and $100\%$, as seen in Fig \ref{tag_density_plot}. Reductions in accuracy occurred primarily in scenarios involving high tag density with tag counts with experiments involving 40 or more tags having lower association accuracy, typically fluctuating between between $80\%$ and $95\%$. Notably, at the three lowest densities, the association accuracy remained near $100\%$ with very few failed associations. At large $R$, the higher tag count caused an increase in asset trajectories that overlapped and trajectories that appeared similar, as seen in Fig \ref{density_trajectories}. Such increases limited the Uncertain Fréchet Distance’s ability to associate tags via path variations. While 100\% accuracy becomes harder to guarantee at scale—even with constant density—consistently exceeding 80\% demonstrates robust theoretical scalability, provided the environment expands proportionally to manage spatial density. 

\begin{figure}
    \centering
    \includegraphics[width=1\linewidth]{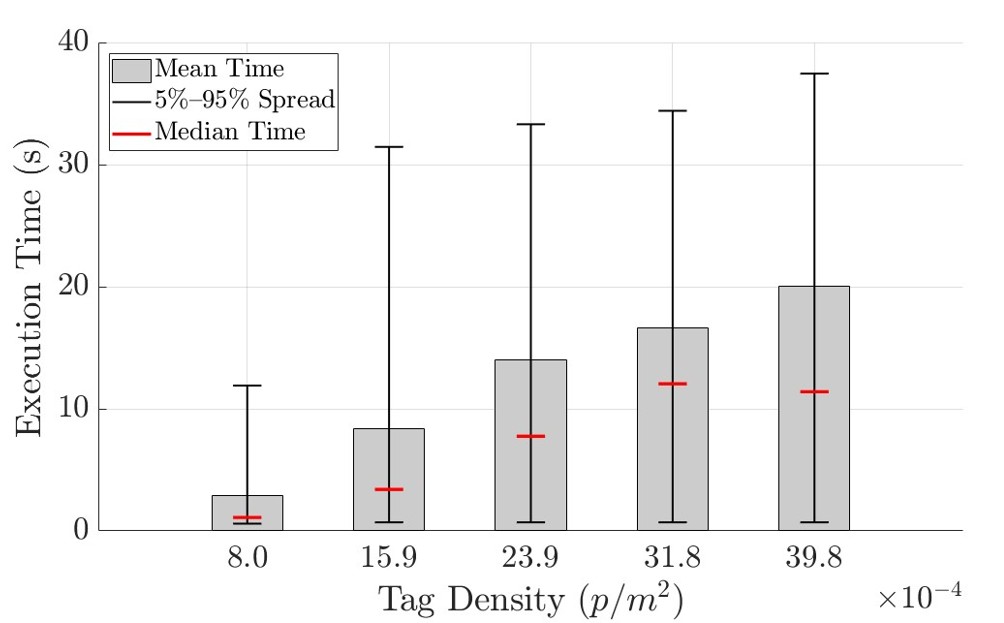}
    \caption{Association time required based on tag density; although the higher tag count increases association time, 95\% of the times this is less than 40s.}
    \label{association_times}
\end{figure}

At lower ranges ($R \leq 40$\,m), the algorithm consistently achieved $100\%$ association accuracy within an observational time frame of $0.6$\,s to $20$\,s. In these scenarios, performance remained independent of tag density, as the models were able to consistently achieve 100\% association accuracy, and required observation times were driven primarily by the total number of tags present. However, as the map radius and tag count increased, association accuracy began to decline while observation times extended to a range of $30$\,s to $40$\,s, as seen in Fig \ref{association_times}.

As shown in Fig.~\ref{tag_density_plot}, alternative methods like the Euclidean norm and DTW performed poorly. The Euclidean ($L_2$) norm relies heavily on spatial separation; while effective for high-precision indoor systems, it fails in outdoor applications where meter-level RFID errors and centimeter-level camera precision cause multiple trajectories to overlap. Unless tags were separated by tens of meters (very low tag densities), Euclidean associations degraded into unreliable associations. Similarly, DTW---which evaluates trajectory shapes by warping temporal sequences---is highly sensitive to signal noise. Although relatively better performing than the Euclidean approach, DTW struggled with overlapping trajectories, rarely exceeding 70\% accuracy for higher densities.

\subsection{Field Experiments}
\subsubsection{Association Results} 
\begin{figure}[htbp]
     \centering
     \begin{subfigure}[b]{0.45\textwidth}
         \centering
         \includegraphics[width=\linewidth]{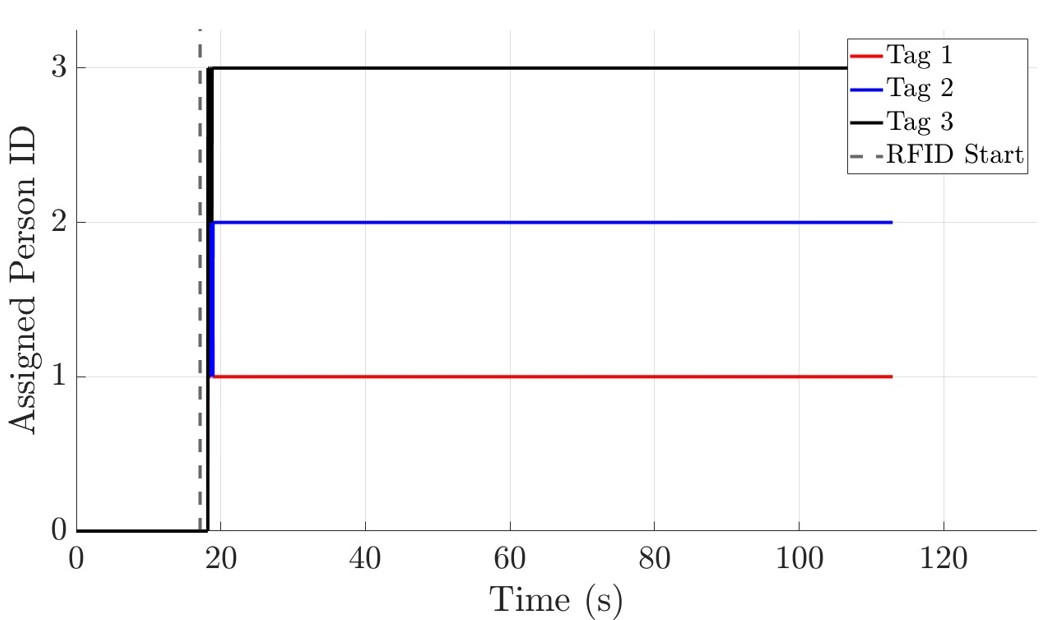}
         \caption{Three People Matching Results}
         \label{three_people_Assoc}
     \end{subfigure}
     \hfill 
     \begin{subfigure}[b]{0.45\textwidth}
         \centering
         \includegraphics[width=\linewidth]{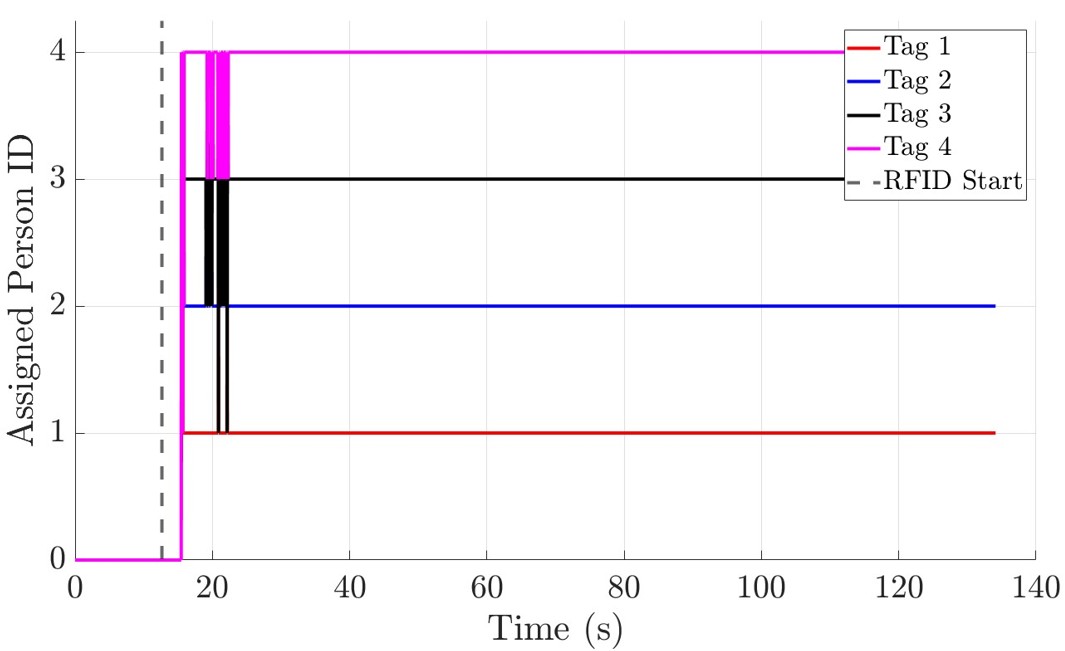}
         \caption{Four People Matching Results}
         \label{four_people_assoc}
     \end{subfigure}
     
     \caption{Labeling accuracy results for real world experiments. The algorithm is capable of quick and accurate association even when the test bed is reduced due to hardware limitations.}
     \label{vid_assoc}
\end{figure}

Fig. \ref{vid_assoc} shows the association results corresponding to three and four tagged person cases moving in the test area. 

\begin{figure}[htbp]
    \centering
    \includegraphics[width=1\linewidth]{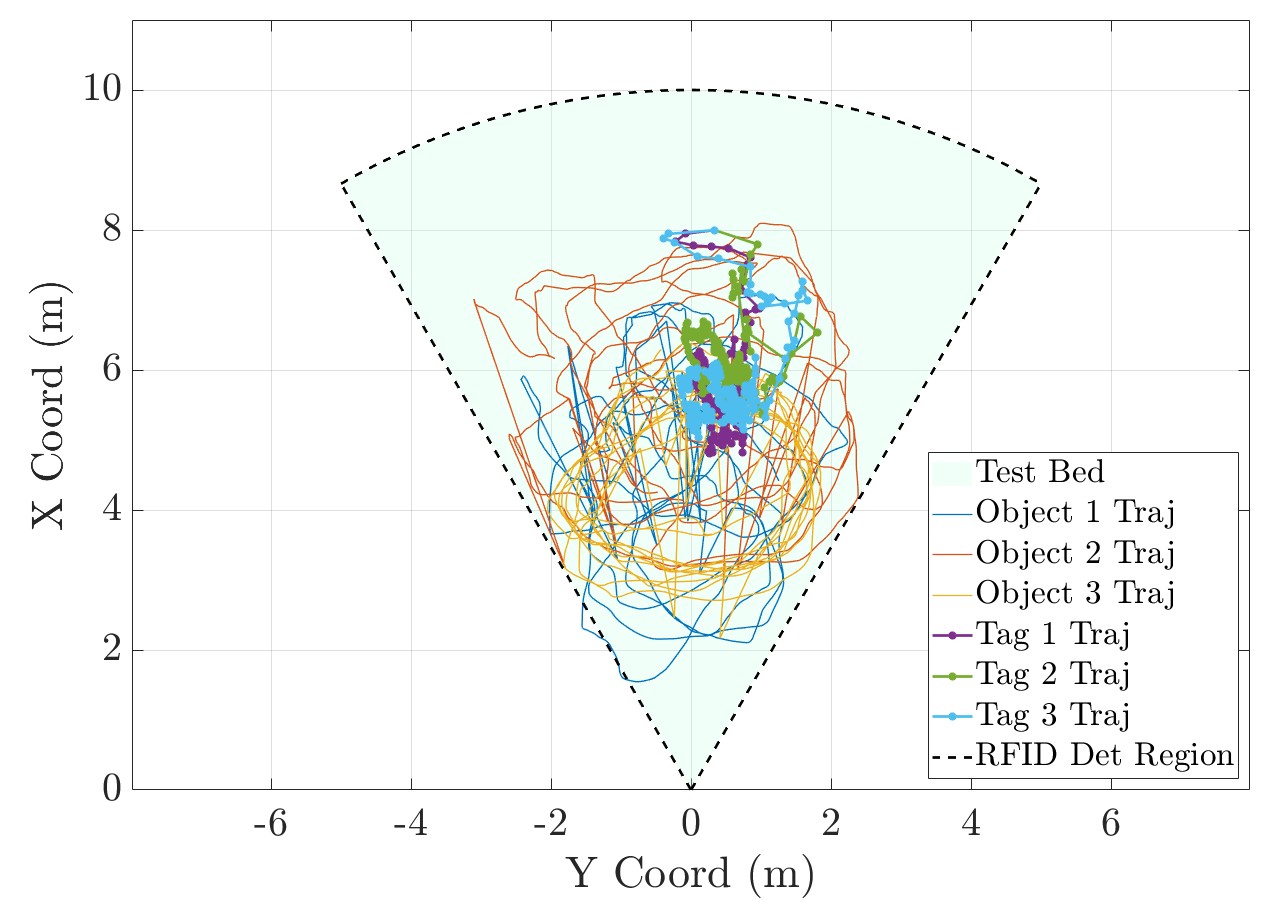}
    \caption{Camera trajectories overlapped with RFID trajectories. Large overlap and scale discrepancies between trajectories hinders association from traditional trajectory matching schemes and requires shape analysis such as using Uncertain Fr\'echet Distance.}
    \label{trajectory_plots}
\end{figure}

In the one- and two-person scenarios, identification occurred in less than a second due to the low density. The three-person scenario required ten RFID measurements per tag (approximately one second of observation) to achieve correct association. The four-person scenario—representing the highest possible density for our test area and hardware—required roughly 70 data points (seven seconds of observation) before the tags were correctly matched. In contrast, the Euclidean norm and DTW failed to produce satisfactory associations regardless of the observation duration. This is likely due to the noise and large discrepancies between the trajectories, as seen in Fig. \ref{trajectory_plots}. 

\subsubsection{Beyond camera Line-of-sight:}
\begin{figure}[htbp]
    \centering
    \includegraphics[width=1\linewidth]{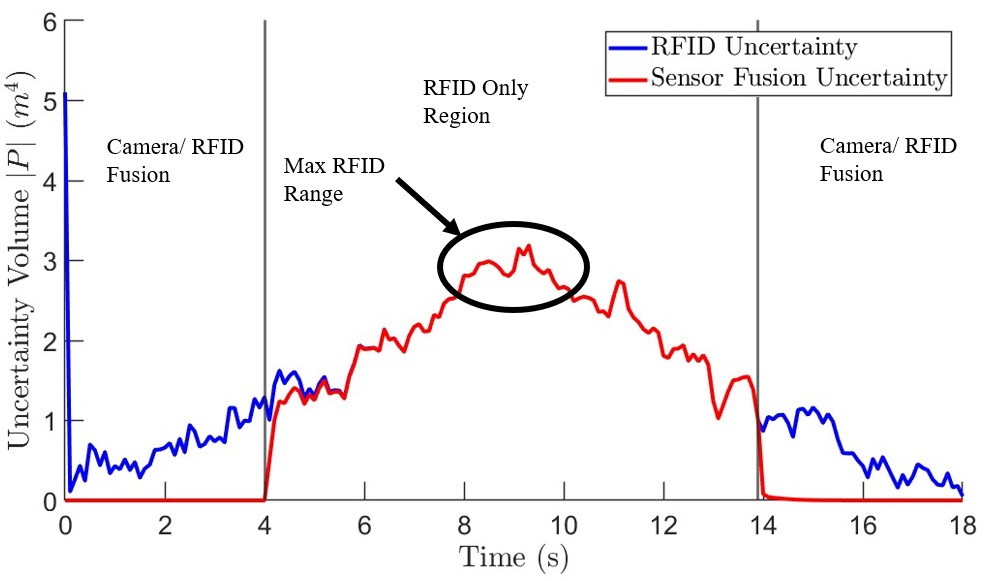}
    \caption{Comparison of generalized variance during sensor transitions. The fusion scheme significantly reduces uncertainty within camera range and leverages EKF state persistence to provide an improved initial trajectory during the transition to RFID-only tracking.}
    \label{uncertainty_volume}
\end{figure}

One of the advantages of the multi-sensor fusion algorithm is the ability to maintain low spatial uncertainty within the camera's line-of-sight while also providing relatively (relative to RFID only) reduced positional uncertainty when the asset transitions into RFID-only tracking beyond the camera's visual range. To evaluate this advantage, experiments were conducted where a participant moved from the camera-monitored region toward the perimeter of the RFID range before returning. Given the consistent nature of the uncertainty responses across trials, a representative result is provided in Fig. \ref{uncertainty_volume}.

During the periods of visual observation, the system leverages standard vision methods to achieve centimeter-level spatial accuracy. This results in a very low uncertainty volume, as the high-precision visual measurements dominate the state update. While the RFID system operates in polar coordinates—leading to increased Cartesian uncertainty as radial distance from the antenna grows—the fusion of both sensors during the overlap period yields significantly higher positional accuracy than either sensor could provide in isolation.

Beyond the $10~m$ range, where the camera can no longer resolve the target, the RFID system becomes the sole source of measurement data. At this transition, the uncertainty region begins to expand. Notably, the superior positional estimate established in the fused phase initially provides the RFID-only tracking with a more accurate state initialization. However, as the system propagates forward in time, the influence of the previous camera data is gradually superseded by the cumulative effect of process noise and the higher variance of the RFID measurements, eventually converging to the standard RFID uncertainty baseline.

\section{Conclusion and Future Work}

This study presents a novel method for enhancing far-range passive RFID localization by associating RFID tags with visually detected objects. The approach combines uncertain Fréchet Distances and Mahalanobis Distance based minimum cost matching to align tag trajectories with visually tracked objects, a strategy not previously explored. The system reliably identifies and associates tagged assets simultaneously, achieving centimeter-level accuracy compared to the sub-meter to meter accuracy typical of standalone RFID systems. This association also enables the Gaussian Process model to impose more reliable kinematic constraints when visual contact is temporarily lost. Results also show that the proposed approach outperforms other popular methods in the literature. However, one of the main limitations of this approach is that performance declines as participant density increases, where spatial overlap among trajectories introduces ambiguity despite trajectory-based matching. Although Fréchet Distances consider path development rather than proximity alone, confined spaces force participants onto similar paths, leading to mis-associations. Hence, it may be necessary to limit the number of tagged assets in the camera's field of view at any given time. The field study was limited to one RFID and camera system; a broader study with various commercial RFID and camera systems may be warranted to validate the performance across all tag densities.

Looking ahead, we are developing optimization functions for robot movement to persistently monitor assets that move in and out of our systems detection range. Through the development of these functions we envision deploying this system with RFID reader(s) and camera(s) mounted on ground-based robots or vehicles operating alongside different forestry and outdoor teams. In such scenarios, personnel and assets—including tools and equipment—would be tagged. This operational scale introduces challenges in coverage planning, persistent monitoring, and scalability, which we identify as key directions for our future research.

\section*{Appendix A}
\label{Appendix_cam}

Depth cameras estimate 3D position at the sub-pixel level though stereo photogrammetry techniques augmented with sophisticated optimization algorithms. Many commercial systems use some variation of methods such as Semi-Global Matching (SGM) to address the correspondence problem between pixels in matching stereo image pairs by bridging the computational efficiency of local matching with the accuracy of 2D global optimization \cite{hirschmuller2008stereo,keselman2017intel}. The theoretical foundation of SGM is the minimization of a global energy function, $E(D)$, which balances the pixel-wise matching cost against a spatial smoothness constraint \cite{hirschmuller2008stereo}:

\begin{equation}
\begin{split}
    E(D) = \sum_{\mathbf{p}} C(\mathbf{p}, D_{\mathbf{p}}) &+ \sum_{\mathbf{q} \in N_{\mathbf{p}}} P_1 T[|D_{\mathbf{p}} - D_{\mathbf{q}}| = 1] \\ 
    &+ \sum_{\mathbf{q} \in N_{\mathbf{p}}} P_2 T[|D_{\mathbf{p}} - D_{\mathbf{q}}| > 1]
\end{split}
\end{equation}

\noindent where $C(\mathbf{p}, D_{\mathbf{p}})$ is the cost of assigning disparity $D$ to pixel $\mathbf{p}$, $N_{\mathbf{p}}$ represents the neighborhood of $\mathbf{p}$, and $P_1$ and $P_2$ are penalty parameters for small and large disparity changes, respectively \cite{hirschmuller2008stereo}. Note that estimating elements $d$ of $D$ is fundamental first step to estimating depth and position.

While the disparity image $D$ can be determined by minimizing $E(D)$, 2D global minimization is NP-complete for many discontinuity-preserving energies \cite{hirschmuller2008stereo}. However, minimization along individual 1D image rows can be performed efficiently, but often suffers from ``streaking'' artifacts when relating 1D optimizations to a full 2D image \cite{hirschmuller2008stereo}. To overcome this, SGM aggregates matching costs from all directions equally, usually between 8 to 16 different paths. The cost along a path $L_{\mathbf{r}}(\mathbf{p}, d)$ in direction $\mathbf{r}$ for pixel $\mathbf{p}$ at disparity $d$ is defined recursively through\cite{hirschmuller2008stereo}:

\begin{equation}
\begin{split}
    L_{\mathbf{r}}(\mathbf{p}, d) = C(\mathbf{p}, d) + \min \begin{cases} 
    L_{\mathbf{r}}(\mathbf{p}-\mathbf{r}, d) \\ 
    L_{\mathbf{r}}(\mathbf{p}-\mathbf{r}, d-1) + P_1 \\ 
    L_{\mathbf{r}}(\mathbf{p}-\mathbf{r}, d+1) + P_1 \\ 
    \min_{i} L_{\mathbf{r}}(\mathbf{p}-\mathbf{r}, i) + P_2 
    \end{cases} \\
    - \min_{k} L_{\mathbf{r}}(\mathbf{p}-\mathbf{r}, k).
\end{split}
\end{equation}

The aggregated costs are then summed across all paths into a single function \cite{hirschmuller2008stereo}:

\begin{equation}
    S(\mathbf{p},d) = \sum_r L_r(\mathbf{p},d)
\end{equation}

By selecting the disparity $d$ that minimizes $S(\mathbf{p},d)$ for each individual pixel, the algorithm creates a dense disparity map. This map is then used to reconstruct the 3D spatial coordinates $(X, Y, Z)$ using standard projective geometry\cite{hartley2003multiple}. The depth $Z$ is inversely proportional to disparity $d$, calculated as $Z = (f \cdot B) / d$, where $f$ is the focal length and $B$ is the baseline between the stereo sensors \cite{hirschmuller2008stereo}.

To efficiently project 2D pixel coordinates $(u, v)$ into 3D space, a $4 \times 4$ reprojection matrix $Q$ is utilized. This transformation relies on intrinsic camera parameters (focal lengths $f_x, f_y$ and principal points $c_x, c_y$), typically obtained through the standard checkerboard calibration technique \cite{zhang2000flexible}. By multiplying the image vector $[u, v, d, 1]^T$ by $Q$, the full 3D point cloud is extracted for asset localization and the calculation of spatial angles such as azimuth \cite{hartley2003multiple}.

\bibliographystyle{IEEEtran} 
\bibliography{references}
\end{document}